\documentclass[11pt,a4paper]{article}
\usepackage[hyperref]{emnlp2020}

\usepackage[utf8]{inputenc} % allow utf-8 input
\usepackage{microtype}
\usepackage{graphicx}
\usepackage{subfigure}
\usepackage{booktabs} 
\usepackage{comment}
\usepackage{hyperref}
\usepackage{mathtools}
\usepackage{color}
\usepackage{graphicx}
\usepackage{CJKutf8}
\usepackage{url}
\usepackage{multicol,multirow}
\usepackage{makecell}
\usepackage{array}
\usepackage{bm}
\usepackage{parskip}
\usepackage[T1]{fontenc}    % use 8-bit T1 fonts
\usepackage{booktabs}       % professional-quality tables
\usepackage{amsmath, amsthm}
\usepackage{amssymb}
\usepackage{amsfonts}

\DeclareMathOperator*{\argmax}{arg\,max}

\usepackage{times}
\usepackage{latexsym}

\usepackage{microtype}
\definecolor{ao}{rgb}{0.0, 0.5, 0.0}
\definecolor{asparagus}{rgb}{0.53, 0.66, 0.42}
\definecolor{amber}{rgb}{1.0, 0.49, 0.0}
\definecolor{alizarin}{rgb}{0.82, 0.1, 0.26}
\definecolor{applegreen}{rgb}{0.55, 0.71, 0.0}
\definecolor{amethyst}{rgb}{0.6, 0.4, 0.8}
\definecolor{auburn}{rgb}{0.43, 0.21, 0.1}
\aclfinalcopy % Uncomment this line for the final submission
\title{Pair the Dots: Jointly Examining Training History  \\ and Test Stimuli for Model Interpretability}

\author{
Yuxian Meng$^\clubsuit$, Chun Fan$^{\spadesuit\bigstar
}$, Zijun Sun$^\clubsuit$, Eduard Hovy$^\blacktriangledown$, Fei Wu$^\blacklozenge$ and Jiwei Li $^{\blacklozenge\clubsuit}$\\
  $^\blacklozenge$Zhejiang University,
  $^\spadesuit$Computer Center of Peking University \\
  $^\bigstar$Peng Cheng Laboratory, $^\blacktriangledown$Carnegie Mellon University\\
  $^\clubsuit$ Shannon.AI\\
  \{yuxian\_meng, zijun\_sun, jiwei\_li\}@shannonai.com\\
  fanchun@pku.edu.cn,
  wufei@zju.edu.cn

}

\date{}

\begin{document}
\maketitle

\begin{abstract}
Any prediction from a model  is made by a combination of  learning history and test stimuli. 
This provides significant insights for improving model interpretability: {\it because of  
which part(s) of
which training example(s), the model attends to which part(s) of a  test example}. 
%Unfortunately, due to the non-differentiable 
%lack of efficiency in algorithms to achieve this goal (e.g.,  not differentiable or extremely time-intensive),  
Unfortunately,
existing methods
to interpret a model's predictions are only able to capture a single aspect of either  test stimuli or  learning history, and  evidences from both are never combined or integrated. 

In this paper, we propose an efficient  and differentiable approach to make it feasible to interpret a model's prediction 
by  jointly examining training history and test stimuli. 
 Test stimuli is first identified  by  gradient-based methods, signifying {\it the part of a test example that the model attends to}. 
The gradient-based saliency scores are then propagated to training examples using  influence functions 
\cite{cook1982residuals,koh2017understanding} to identify 
{\it which part(s) of
which training example(s)} 
make the model attends to the test stimuli. 
The system is differentiable and time efficient:  
the adoption of saliency scores from gradient-based methods 
allows us to efficiently 
trace a model's prediction through test stimuli, and then back to training examples through influence functions.

We demonstrate that the proposed methodology  offers clear
explanations about neural model decisions, along with being useful for performing error analysis, crafting adversarial examples
 and fixing  erroneously classified examples. 
\end{abstract}

\section{Introduction} 
While
neural models match or outperform the performance of  state-of-the-art systems on a variety of  tasks,
they intrinsically suffer a severe shortcoming of being 
 hard to explain \citep{simonyan2013deep,bach2015lrp,montavon2017explaining,kindermans2017learning}.
Unlike traditional feature-based classifiers that assign weights to human-interpretable features, 
neural network models operate like a black box with multiple layers non-linear operations on input representations \citep{glorot2011deep,kaiming2016resnet,vaswani2017transformer}. 
The lack of interpretability not only significantly limits the scope of its applications in areas where model interpretations are essential, but also makes it hard to perform model behavior analysis and error analysis.
%especially when the model is extremely fragile, e.g., when adversarial examples are present \citep{goodfellow2014explaining,kurakin2016adversarial,papernot2017practical,athalye2018synthesizing,jia2017adversarial,zhou2020defense}. 

Any prediction from a model  is made by a combination of  learning history and test stimuli. 
To explain the prediction of a model, we need to answer the following question, 
{\it because of  
which part(s) of
which training example(s), the model attends to which part(s) of a  test example}. 
Unfortunately, existing strategies, 
test-focused methods \cite{simonyan2013deep,li2015visualizing,li2016understanding,tenney2019bert,clark2019does,wallace2019allennlp}, and 
training-focused methods \cite{koh2017understanding,NIPS2018_7849,barham2019interpretable}, only capture a single aspect, either test stimuli
or learning history:
Training-focused methods
 center on 
detecting the salient part of a test example that a model attends to, but are incapable of 
  explaining {\it why} the model looks at the identified part.
Training-focused methods identify  influential training examples that hold responsible for 
 a model's prediction, but are incapable of 
  explaining {\it how} the identified training examples affects a prediction. 
We thus need a model that can 
jointly examine training history and test stimuli. 

The difficulty that hinders joint modeling lies at the algorithm level, where models for 
examining  test stimuli are usually not differentiable with respect to the training examples, and that measuring the influence of a training example can also be  time-intensive as it requires retraining the model. 
To overcome  these difficulties, in this paper, we propose a differentiable model that can straightforwardly traces the prediction though test stimuli, and backs to training examples: 
Test stimuli is first identified  using  gradient-based methods, signifying {\it the part of a test example that the model attends to}. 
Next, the gradient-based saliency scores are propagated to training examples using  influence functions 
\cite{cook1982residuals,koh2017understanding} to identify 
{\it which part(s) of
which training example(s)} 
make the model attends to the test stimuli. 
The key advantages from the proposed system come in two folds: (1) 
the system is differentiable: 
the adoption of saliency scores from gradient-based methods 
are differentiable with respect to training examples 
through the influence function; 
(2) the system is time-efficient and the training model does not need to be retrained to examine learning history. 
In this way, we are able to efficiently 
trace a model's prediction through test stimuli, and then back to training examples. 
%, e.g.,  the part of an input example that the model attends to, or contributes negatively or positively to a model's decision. 

\begin{figure}[t]
    \centering
    \includegraphics[scale=0.7]{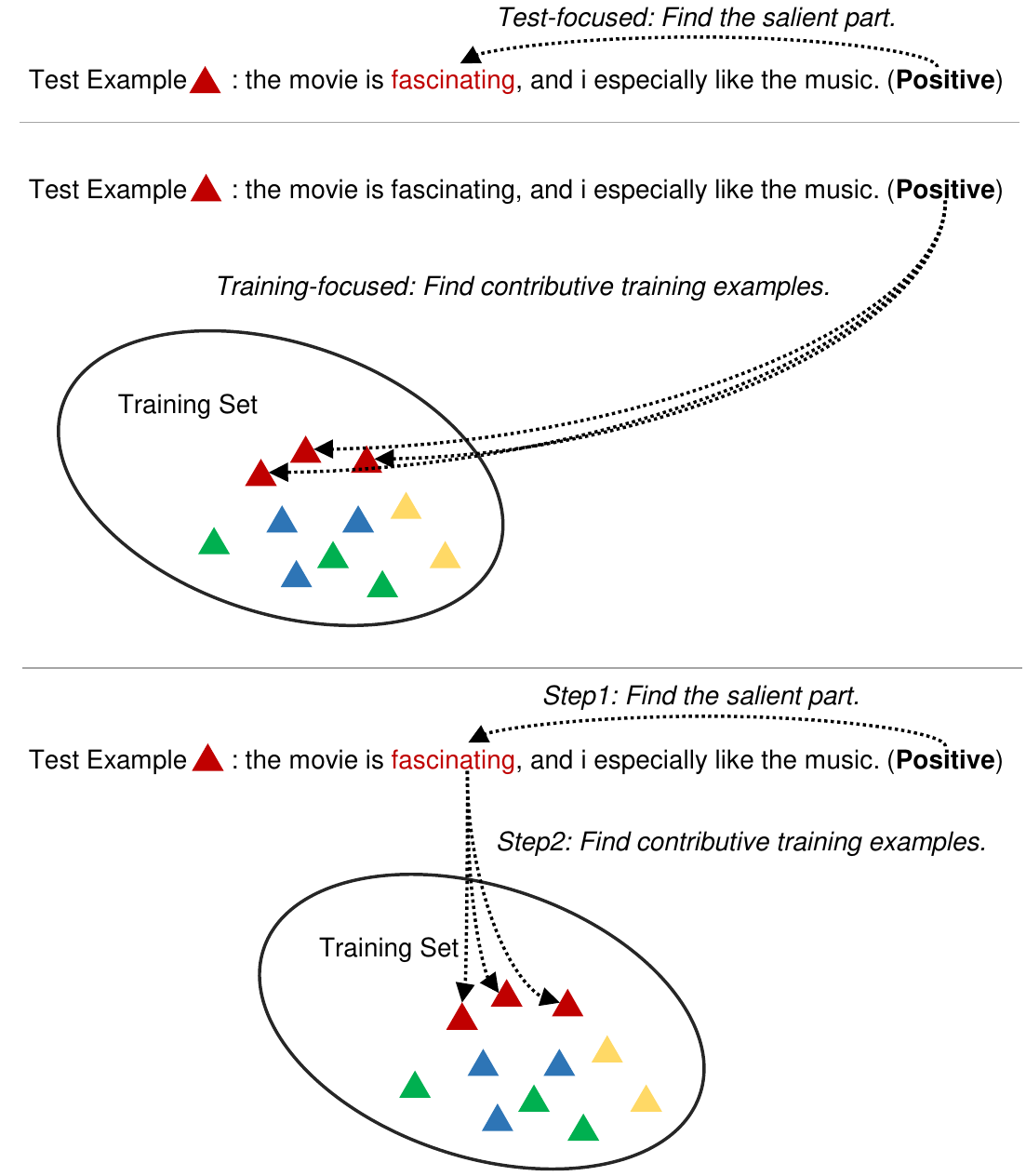}
    \caption{An illustration of different methods for interpreting neural decisions. {\it Top}: test-focused methods aim at finding the salient part of a given test example. {\it Middle}: training-focused methods associate training examples responsible for predicting a given test example. {\it Bottom}: the combination of both test-focused and training-focused methods extend beyond each of them, thereby giving more comprehensive explanations.}
    \label{fig1}
\end{figure}

In summary, the contributions of this paper are as follows: 
(1) we propose an efficient and differentiable model that 
jointly examines training history and test stimuli for model interpretability;
 (2)  we show that the proposed methodology is a versatile  tool for a wide range of applications, including error analysis, crafting adversarial examples
 and fixing  erroneously classified examples.

The rest of this work is organized as follows: in Section \ref{related-work} we detail related work of test-focused and training-focused methods for model interpretations.
The proposed model is described in Section 3. Experimental results are shown in Section 4, followed by a brief conclusion in Section 5.

\section{Related Work} 
\label{related-work}
Test-focused interpretation
methods
 aim at  finding  the most salient input features responsible for a prediction: 
 %There have been much work on test-focused interpretation methods. 
\citet{adler2018auditing,datta2016algorithmic} explained neural models by perturbing various parts of the input, and compared the performance change in downstream tasks to measure feature importance;
\citet{ribeiro2016i} proposed LIME, 
which 
interprets  model predictions by  approximating model predictions; 
\citet{shrikumar2017learning} proposed DeepLIFT, which assigns each input a value to represent  importance.  
\citet{10.1371/journal.pone.0130140} used  Layer-Wise Relevance Propagation (LRP) to explain which pixels of an image are relevant for obtaining a model's decision;  
\citet{alvarez2017causal} presented a causal framework to explain structured models;  
Other methods include directly visualizing input features by computing saliency scores with gradients \citep{simonyan2013deep,srinivas2019full}, 
using a surrogate model that learns to estimate the degree to which certain inputs cause outputs \citep{schwab2019cxplain,zhang2018unsupervised},
and selecting  features that cause cell activations \citep{zeiler2014visualizing,springenberg2014striving,kindermans2017learning} or 
 signals  through layers
 \citep{montavon2017explaining,sundararajan2017axiomatic,lundberg2017unified}.

For training-focused methods,  \citet{koh2017understanding} used influence functions \citep{cook1982residuals,cook1986assessment}  to  identify training points that are most responsible for a given prediction. 
%The proposed framework can efficiently detect how a slight change in a specific training data point affects the performance of a given test example, by approximating derivatives of the loss function with respect to small perturbations. 
A followup work from \citet{han2020explaining} finds
that influence functions are particularly useful for tasks like natural language inference

In the context of NLP, 
%except for the above mentioned gradient-based methods \citep{li2015visualizing} and perturbation-based methods \citep{li2016understanding,feng2018pathologies}, other methods for interpretation are proposed, including 
methods for model interpretation  include
 generating justifications to form  interpretable summaries \citep{lei2016rationalizing}, studying the roles of vector dimensions \citep{shi-etal-2016-neural}, and  analyzing  hidden state dynamics in recurrent neural networks \citep{hermans2013training,karpathy2015visualizing,greff2016lstm,strobelt2016visual,kadar2017representation}.
The attention mechanism has been used for model interpretation \cite{jain2019attention,serrano2019attention,wiegreffe-pinter-2019-attention,vashishth2019attention}. 
%Attention-based methods plot the attention heatmaps in terms of each input token (or feature) with respect to the output to see which part of the input the model attends to most, meaning it may be the most crucial part for prediction. 
For example, \citet{ghaeini2018interpreting} applied the attention mechanism to interpret models for the task of natural language inference;
\citet{vig2019analyzing,tenney2019bert,clark2019does} analyzed the attention structures inside self-attentions, and found that linguistic patterns
emerge in different heads and layers.

A less relevant line of work is model poisoning, which  manipulates the training process using poisoned data \citep{biggio2012poisoning,mei2015using,munoz2017towards,NIPS2018_7849},
%Similar to \citet{koh2017understanding}, these methods also target at perturbing training points, which can be in some degree regarded as a way of interpreting how neural models can be affected by training examples.
or 
adversarial examples \citep{goodfellow2014explaining,kurakin2016adversarial,moosavi2016deepfool,madry2017towards,papernot2017practical,athalye2018synthesizing}.
It has been  proven that  training neural models with the combination of  clean data and adversarial data can not only benefit  test accuracy \citep{ebrahimi2017hotflip,jia2019certified,wang2020fast,Zhu2020FreeLB,zhou2020defense}, but also 
offers
explanations  on the vulnerability of 
 neural models  \citep{miyato2016adversarial,papernot2016crafting,zhao2017generating,ijcai2018-601,barham2019interpretable}.

\section{Approach}
\label{approach}
Consider a task that maps an input $\mathbf{x}$  to a label $y$, where $\mathbf{x}$ 
is a list of tokens $\mathbf{x} = \{x_1,x_2,...,x_N\}$ and $L$ is the length of $\mathbf{x}$. 
Each token $x_t \in \mathbf{x}$ is associated with a
$d$-dimensional 
 word vector $\mathbf{e}_i$. 
The concatenation of $\mathbf{e}_t$ forms
the $d\times L$ dimensional input embedding
 $\mathbf{E}$. 
A model takes as input $\mathbf{E}$, and 
 maps it 
 to single vector $\mathbf{h}_x$.
 $\mathbf{h}_x$ is next projected to 
   a distribution over labels $\mathbf{y}$ using the softmax operator.\footnote{Here we use the widely used neural framework that first maps input $\mathbf{E}$ to $\mathbf{h}_x$ and then maps $\mathbf{h}_x$ to the label $y$ for illustration purposes. It is worth noting that the proposed framework is a general one and can be easily adapted to other setups.}
In the following parts, we first introduce how  to identify the salient feature of a test example,and then describe how to find 
training examples that are 
the most contributive  to the selected salient token(s).

\subsection{Detecting the Salient Part of  a Test Example}
\label{test}
To answer the question of {\it the model attends to which part(s) of a given test example},
we need to first decide which 
unit(s) of $\mathbf{E}$ make(s) the most
significant contribution to $S_y(\mathbf{E})$, which is the probability of assigning label $y$ to $x$. 
Different widely used 
test-focused visualization 
methods can serve this purpose, such as 
LRP \cite{10.1371/journal.pone.0130140}, LIME \cite{ribeiro2016i}, DeepLIFT \cite{shrikumar2017learning}.
But they can not serve our purpose since they are not differentiable with respect to training examples. 
We thus propose to use the gradient method \cite{simonyan2013deep,li2015visualizing} because of the key advantage that it is differentiable and  can
thus
 be straightforwardly combined with the influence function to identify 
contributive  training examples, as will be illustrated in the next subsection. 

For the gradient-based  model, 
since $S_y(\mathbf{E})$ is a highly non-linear function with respect to $\mathbf{E}$, 
we approximate it
 with a linear function of $\mathbf{e}$ by computing the first-order Taylor expansion:
\begin{equation}
S_y(\mathbf{e})\approx w_y(\mathbf{e})^\top \mathbf{e}+b
\end{equation}
where $w_y(\mathbf{e})\in \mathbb{R}^{d\times 1}$ is the derivative of $S_y(\mathbf{E})$ with respect to the embedding $\mathbf{e}$:
\begin{equation}
w_y(\mathbf{e})=\nabla_\mathbf{e} S_y(\mathbf{E})
\end{equation}
$w_y(\mathbf{e})$
the sensitiveness of  the final decision to the change in one particular dimension, 
telling us how much word embedding dimensions contribute to the final decision.
The saliency score is given by $w_y(\mathbf{e})$. 
Once we have the saliency score for each token embedding $\mathbf{e}_i$, we can accordingly select the salient tokens:
 we first sum up  values of constituent dimensions in the word embedding and using $s(x_t)=\sum_j w_y(\mathbf{e}_t)_j$ as the saliency score for the  token $x_t$.
%, and then we choose the top 1\% the final salient tokens that are then used to select training examples.
%Here we need a hard criteria to decide whether a token is salient or not. We use the token erasing strategy described in \cite{li2016understanding,ren2019generating}, where we say
%that $x_t$ is salient if replacing it with an UNK token flips the label of $x$. 
Next we can rank all constituent tokens $x_t$ by saliency scores to identify the token that a model attends to. 

\subsection{Training Examples' Influence on the Detected Salient Parts}
For each of the selected salient tokens $x_t$, we need to answer the question {\it because of  
which part(s) of
which training example(s), the model attends to $x_t$}.
 Here 
  we use the method of influence functions \citep{cook1982residuals,koh2017understanding} for this purpose. The key idea of 
  the influence function is to measure 
  the change of the model's parameters when a training point changes. 
Since the saliency score
described in the previous subsection 
 can be viewed as a function of the model's parameters, 
the influence of changing of a training point on the saliency score can be straightforwardly computed through the chain rule.

Specifically, given training points $\{(\mathbf{x}_i,y_i)\}$, let $\theta\in \mathbb{R}^{P}$ be  model parameters and $L(\mathbf{x}_i, y_i;\theta)$ be the loss, which  is negative log likelihood.
 Let $z_i = (\mathbf{x}_i, y_i)$ for  simplification. 
  $\theta$  is learned by minimizing the loss given as follows: 
 \begin{equation}
 {\theta} = \arg\min_\theta\frac{1}{n}\sum_{z_i} L(z_i ;\theta)
 \end{equation}
The importance of $z_i$ is measured by 
 the change of $\theta$ when $z_i$ is removed from the training set, given by:
 ${\theta}_{-z}-{\theta}$, where ${\theta}_{-z}\triangleq\arg\min_\theta\sum_{z_i: z_i\not=z}L(z_i;\theta)$. 
But this is extremely computationally intensive  since it requires retraining the model on the training set with $z_i$ removed. 
Fortunately, we can use influence functions \cite{cook1982residuals,koh2017understanding} to efficiently approximate the value.  We define ${\theta}_{z,\epsilon}=\arg\min_\theta\frac{1}{n}\sum_{i=1}^nL(z_i;\theta)+\epsilon L(z;\theta)$ to be the learned parameters when the loss function is upweighed by some small change $\epsilon$ on a training point $z$, and thus ${\theta}_{-z}$ is exactly ${\theta}_{z,-1}$. By leaving the proof to Appendix \ref{appendixA}, the influence of upweighting $z$ on $\theta$ (denoted by $I(z,{\theta})$)
      is given as follows:
\begin{equation}
  I(z,{\theta})=\frac{\mathrm{d}{\theta}_{z,\epsilon}}{\mathrm{d}\epsilon}\Big|_{\epsilon=0}=-H^{-1}_{{\theta}}\nabla_\theta L(z;{\theta})
  \label{eq3}
\end{equation}
where $H_{{\theta}}=\frac{1}{n}\sum_{z_i}\nabla^2_\theta L(z_i;{\theta})$ is the Hessian matrix.
$H_{{\theta}}\in \mathbb{R}^{P\times P}$, where $\nabla^2_\theta L(z_i;{\theta}) = \frac{\partial^2 L(z_i;{\theta})}{\partial\theta_i\partial\theta_j}$. 
 Using linear approximations, we can immediately get ${\theta}_{-z}-{\theta}\approx-\frac{1}{n} I(z,{\theta})$.

 Our goal is to find the influence of a training example  $z$ on a salient token $x_t$, i.e., the saliency score $w_y(\mathbf{e})$.
 Since $w_y(\mathbf{e})$  is a function of parameters ${\theta}$,
we can apply the chain rule to measure the influence  of $z_i$ on $w_y(\mathbf{e})$, denoted by $I(z,w_y(\mathbf{e}))$: 
\begin{equation}
  \begin{aligned}
  I(z,w_y(\mathbf{e}))&\triangleq\frac{\mathrm{d}w_y(\mathbf{e};{\theta}_{z,\epsilon})}{\mathrm{d}\epsilon}\Big|_{\epsilon=0}\\
  &=\frac{\mathrm{d}w_y(\mathbf{e};{\theta}_{z,\epsilon})}{\mathrm{d}{\theta}_{z,\epsilon}}\frac{\mathrm{d}{\theta}_{z,\epsilon}}{\mathrm{d}\epsilon}\Big|_{\epsilon=0}\\
  &=-\nabla_\theta w_y(\mathbf{e})^\top H^{-1}_{{\theta}}\nabla_\theta L(z;{\theta})
  \label{eq4}
  \end{aligned}
\end{equation}
Now for a given salient token $x_t$ with its word embedding $\mathbf{e}_t$, we are able to quantitatively measure the contribution of each training point $z_i$ on the detected salient part of the test exampled, 
quantified by $I(z_i,w_y(\mathbf{e}))$. 

\paragraph{Perturbing a Training Point} The above process describes how a  training point can influence the model parameters.
Further, we would like to measure how a {\it perturbed} training point can affect the model.
Due to the discrete nature of NLP, an input example
 $\mathbf{x} = \{x_1,x_2, ..., x_t, ..., x_N\}$ will be perturbed to $\tilde{\mathbf{x}}= \{x_1,x_2, ..., x_t', ..., x_N\}$, with its constituent token $x_t$ altered to $\tilde{x}_t$. 
 Concretely, for a training point $z=(\mathbf{x}, y)$,  we define its perturbed counterpart 
 $\tilde{z}=(\tilde{\mathbf{x}}, y)$.
 Let ${\theta}_{z, \tilde{z} ,\epsilon}\triangleq\arg\min_\theta\frac{1}{n}\sum_{i=1}^nL(z_i;\theta)+\epsilon L(\tilde{z};\theta)-\epsilon L(z;\theta)$ be the parameters after substituting $z$ with $\tilde{z}$. Similar to Eq.\ref{eq3}, the influence of changing $z$ to  $\tilde{z}$ is given as follows: 
\begin{equation}
  \begin{aligned}
    I&(z,\tilde{z}, {\theta})\triangleq\frac{\mathrm{d}{\theta}_{z, \tilde{z},\epsilon}}{\mathrm{d}\epsilon}\Big|_{\epsilon=0}
    = I(\tilde{z},{\theta}) - I(z,{\theta}) \\
    &=-H^{-1}_{{\theta}}(\nabla_\theta L(\tilde{z};{\theta})-\nabla_\theta L(z;{\theta}))
  \end{aligned}
\end{equation}
Again, by using the chain rule, we can measure the influence of changing $z$ to $\tilde{z}$ on the saliency score $w_y(\mathbf{e})$: 
\begin{equation}
  \begin{aligned}
    &I(z,\tilde{z}, w_y(\mathbf{e}) )\triangleq\frac{\mathrm{d}w_y(\mathbf{e};{\theta}_{z,\tilde{z},\epsilon})}{\mathrm{d}\epsilon}\Big|_{\epsilon=0} \\
    & = I(\tilde{z},w_y(\mathbf{e})) - I(z,w_y(\mathbf{e})) \\
    &=-\nabla_\theta w_y(\mathbf{e})^\top H^{-1}_{{\theta}}(\nabla_\theta L(\tilde{z};{\theta})-\nabla_\theta L(z;{\theta}))
    \label{eq5}
  \end{aligned}
\end{equation}
To this end, we are able to quantitively compute the influence of perturbing a training example $z$ to $\tilde{z}$ on the salient part of a test example. 

It is also interesting to look at a special case of Eq.\ref{eq5}, where $\mathbf{x} = \{x_1,x_2, ..., x_t, ..., x_N\}$ is perturbed to $\tilde{\mathbf{x}}(t, \text{UNK})= \{x_1,x_2, ..., \text{UNK}, ..., x_N\}$,
with ${x}_t=\text{UNK}$, and UNK is the special unknown word token.\footnote{This purpose can also be achieved by replacing the token embedding with an all-zero vector.}
$I(z, \tilde{z}(t, \text{UNK}),w_y(\mathbf{e}))$ actually measures the change on the saliency score when a certain word $x_t$ is erased, which is equivalent to the influence of the individual word $x_t$, denote by $   I(x_t, w_y(\mathbf{e}) )$:
\begin{equation}
    I(x_t, w_y(\mathbf{e}) ) = I(z,w_y(\mathbf{e})) -  I(\tilde{z}(t, \text{UNK}),w_y(\mathbf{e}))\\
    \label{token}
\end{equation}

\section{Applications}
The proposed methodology facilitates the following important use cases, including
understanding model behavior and
 performing error analysis, 
 generating adversarial examples,
 and fixing erroneously classified examples. 
We use the following widely used benchmarks  to perform analysis:

(1) the 
Stanford Sentiment Treebank (SST), a widely used  benchmark 
for
 neural model evaluations.
The task is to perform both
fine-grained (very positive, positive, neutral, negative and very negative) and coarse-grained (positive
and negative) classification at both the phrase and
sentence level. 
For more details about the dataset,
please refer to \cite{socher2013recursive}. 

(2) the IMDB dataset \citep{maas2011learning}, which consists of 25,000 training samples and 25,000
test samples, each of which is labeled as positive or negative sentiment. 

(3)  the
AG’s News dataset, which is a collection  news articles  categorized into four classes: World, Sports, Business
and Sci/Tech. Each class contains 30,000 training
samples and 1,900 testing samples.

\begin{table*}[t]
\centering
\small
\begin{tabular}{p{7.5cm}p{7.5cm}}
\toprule
{\bf TreeLSTM}  & {\bf BERT} \\\midrule
\multicolumn{2}{l}{{\it Test example 1}:  i  {\color{red}{\bf loved}} it ! } \\
(a) if you 're a fan of the series you 'll  {\color{ao} {\bf love}} it.  &  (a) i  {\color{ao} {\bf  loved}} this film . \\
(b) old people will  {\color{ao} {\bf  love}} this movie. & (b) you 'll probably  {\color{ao} {\bf  love}} it .\\
(c) ken russell would  {\color{ao} {\bf love}} this .&(c)  a movie i  {\color{ao} {\bf loved}} on first sight \\
(d) i  {\color{ao} {\bf loved}} this film . & (d) old people will  {\color{ao} {\bf love}} this movie. \\
(e) an ideal  {\color{ao} {\bf love}} story for those intolerant of the more common saccharine genre . & (e) i  {\color{ao} {\bf like}} this movie a lot ! \\\midrule
\multicolumn{2}{l}{{\it Test example 2}: 
it 's not life-affirming -- its vulgar and mean , but i  {\color{red}{\bf liked}} it .} \\
(a) i {\color{ao} {\bf like}} it . & (a) as an introduction to the man 's theories and influence , derrida is all but useless ; as a portrait of the artist as an endlessly inquisitive old man , however , it 's {\color{ao} {\bf invaluable}}  \\
(b) the more you think about the movie , the more you will probably {\color{ao} {\bf like }}it . &(b) i {\color{ao} {\bf like}} it . \\
(c) one of the {\color{ao} {\bf best}} , most understated performances of  jack nicholson 's career &(c) i liked the movie , but i know i would have liked it more if it had just gone that one step further .\\
(d) it 's not nearly as fresh or {\color{ao} {\bf enjoyable}} as its predecessor , but there are enough high points to keep this from being a complete waste of time . &(d)sillier , cuter , and shorter than the first  as best i remember , but still a very {\color{ao} {\bf good}} time at the cinema \\
(e)  i {\color{ao} {\bf liked}} it just enough . & (e) too daft by half ... but supremely {\color{ao} {\bf good}} natured . \\\bottomrule
\end{tabular}
\caption{
The most salient token in the test example regarding the golden label $y$ is labeled in {\color{red}{\bf red}}.
Each test example is paired with top 5 training examples that are the most responsible for the salient region  by $ I(z,w_y(\mathbf{e}))$ in Eq.\ref{eq4}. 
For each extracted training example, its constituent token that is  the most responsible for the salient region by $ I(x_t, w_y(\mathbf{e}) )$ in Eq.\ref{token} is marked in {\color{ao} {\bf green}}.}
\label{behavior}
\end{table*}

\subsection{Model Behavior Analysis and Error Analysis}
\paragraph{Model Behavior Analysis}
The proposed paradigm provides a direct way to perform model behavior analysis: 
for a test example with 
the gold
label $y$, we can first identify the salient region that the model   focuses on based on $\nabla_\theta w_y(\mathbf{e})$.
Next, we can identify
the most contributive 
 training examples  for the detected salient region based on $I(z,w_y(\mathbf{e}))$ in Eq.\ref{eq4}.
The influence within individual words of training examples 
can be measured by $ I(x_t, w_y(\mathbf{e}) )$ in 
 Eq.\ref{token}. 

Examples from SST are present at
Table \ref{behavior}. 
We use two models as the backbone, TreeLSTMs \cite{tai2015improved,li-etal-2015-tree}  and BERT \cite{devlin2018bert}. 
For the relatively easy test
 example {\it i loved it}, 
the model straightforwardly identifies the sentiment-indicative token {\it loved} within the  test example based on  $\nabla_\theta w_y(\mathbf{e})$. 
Further, based on Eq.\ref{eq4} and \ref{token}, we can identify  training examples contributive to
the salient 
 {\it loved} in the test example, 
which are 
mostly positive training points that directly contain the keyword {\it love} or {\it loved}. 
Interestingly, one responsible training point (e) from TreeLSTMs is confused about the multi-senses for the word {\it love}: {\it love} in {\it love story} and {\it love} in {\it i love it}.
In contrast, BERT does not make similar mistakes, which explains its better performances. 
For the second example {\it it 's not life-affirming -- its vulgar and mean , but i liked it}, a concessive  sentence  with a contrast conjunction,  training points that hold responsible  involve both 
simple sentences with the mention of the keyword (e.g.,  {\it  i like it .} or {\it i liked it just enough .}) and 
 and concessive sentences that share the sentiment label.

\begin{table*}[!ht]
\small
\center
\begin{tabular}{p{7.5cm}p{7.5cm}}\\\toprule
{\bf Incorrect label} $y'$ for $\nabla_\theta w_{y'}(\mathbf{e})$ &
{\bf Correct label} $y$ for $\nabla_\theta w_y(\mathbf{e})$ \\
the {\color{red}{\bf  best}} way to hope for any chance of enjoying this film is by lowering your expectations.  &
 the best way to hope for any chance of enjoying this film is by {\color{ao} {\bf lowering}} your expectations.   \\\midrule
(a) mr. deeds is , as comedy goes , very silly -- and in the {\color{amethyst} {\bf best}} way . & (a) many     {\color{auburn} {\bf  shallower}} movies these days seem too long , but this one is egregiously short .
 \\
(b) a terrific b movie -- in fact , the {\color{amethyst} {\bf best}} in recent memory. &   (b) below is well   {\color{auburn} {\bf  below}} expectations .
\\ 
(c) one of the {\color{amethyst} {\bf greatest}} films i 've ever seen .  &   (c) you 'll get the enjoyable basic   {\color{auburn} {\bf  minimum}}.
\\
(d) another {\color{amethyst} {\bf best}} of the year selection .  & (d) low comedy does n't come much   {\color{auburn} {\bf  lower}} . \\
(e) the {\color{amethyst} {\bf best}} way to describe it is as a cross between paul thomas anderson 's magnolia and david lynch 's mulholland dr. &(e) you 'll   {\color{auburn} {\bf  forget}} about it by monday , though , and if they 're old enough to have dev     eloped some taste , so will your kids .
\\\bottomrule
\end{tabular}
\caption{Error analysis for an erroneously classified  
 example: for a test example  erroneously classified as $y'$, the token in {\color{red}{\bf  red}} denotes the salient word with respect to $y'$, and tokens in  {\color{amethyst} {\bf purple}} denote the top responsible tokens in the top influential training examples.  
For contrasting purposes, salient test token and corresponding training examples for the correct label $y$
are listed 
on the right hand side:
the token in {\color{ao}{\bf  green}} denotes the salient word with respect to the correct label $y$, and tokens in  {\color{auburn} {\bf brown}} denote the top responsible tokens in the top influential training examples. }
\label{error}
\end{table*}

\paragraph{Error Analysis}
We can make minor changes for the purpose of performing error analysis: 
for a test example with erroneously labeled as $y'$, we can first identify the salient region that the model   focuses on based on $\nabla_\theta w_{y'}(\mathbf{e})$, which the region that the model should not focus on. 
Next, we can identify training examples that are responsible for the  salient region by $I(z,w_{y'}(\mathbf{e}))$. 
Examples are shown in Table \ref{error}:
for the input text {\it the best way to hope for any chance of enjoying this film is by lowering your expectations}, 
the model inappropriately focuses on the positive word {\it best},  
leading to an incorrect prediction. 
Training examples that hold responsible for detecting the salient word 
involve {\it in fact , the best in recent memory}, and {\it  another best of the year selection}, both of which contain the keyword {\it best}, but are of different meanings. 
We can see that the model cannot fully  disambiguate between the positive sentiment that the word {\it best} holds 
in {\it another best of the year selection}
 and the neutral sentiment it holds in the context of {\it the best way}, leading to the model focusing on the region of a test example that it should not focus on, resulting in the final incorrect decision. 

\begin{table*}[!ht]
\center
\small
\begin{tabular}{p{2cm}p{13cm}}\toprule
\makecell[c]{{\it Test Example}\\ {$\hat{y}$=Positive} \\$p(\hat{y})=0.782$  }& 
\multicolumn{1}{m{13cm}}{Having watched this film years ago, it never faded from my memory. I always thought this was the {\color{red}{\bf finest}} performance by Michelle Pfeiffer that I've seen. But, I am astounded by the number of {\color{ao}{\bf negative}} reviews that this film has received. After seeing it once more today, I still think it is powerful, moving and couldn't care less if it is "based loosely on King Lear".I now realize that this is the {\color{red} {\bf greatest}} performance by Jessica Lange that I've ever seen - and she has had accolades for much shallower efforts. A Thousand Acres is complex, human, vibrant and immensely moving, but surely doesn't present either of the primary female leads with any touch of glamour or "sexiness"...
Perhaps one reason for this film's  {\color{ao} {\bf underwhelming}} response lies in the fact that the writer (Jane Smiley(, screenplay (Laura Jones), and director (Jocelhyn Moorehouse) are all women. I know that, in my younger days, I wouldn't have read a book written by a woman. I didn't focus on this fact until years later.If you haven't seen this movie or gave it a chance in the past, try watching it anew. Maybe you are ready for it. }\\\midrule
\makecell[c]{{\it Training Example} \\  Golden $\hat{y}$}&
\multicolumn{1}{m{13cm}}{...StarDust would make an unexpected twist and involve you more and more into the story.the actors are great - even the smallest part is performed with such talent it fills me with awe for the creators of this movie. Robert De Niro is gorgeous and performs with such energy that he simply steals the show in each scene he's in. Michelle Pfeiffer is the {\color{red}{\bf perfect}} ({\color{amethyst}{\bf flawless}}) witch, and Claire Danes a wonderful choice for the innocent and loving 'star', Yvaine. Other big names make outstanding roles. I had the filling everyone is trying to give his best for this movie. But once again, the story by Neil Gaiman, all the little things he 'invented' for this universe - simply outstanding...}\\\midrule
\makecell[c]{{\it Training Example}\\ Incorrect $y'$}&
\multicolumn{1}{m{13cm}}{The show is really funny. Nice theme. Jokes and one liners are really good. With little extra tuning it can become a very popular show. But the only major {\color{ao}{\bf negative}} ({\color{amethyst} {\bf unfavorable}}) point of this show is the cast. David Spade does a great job as Russell, Megyn Price does a good job. 
But who the hell did cast Patrick Warburton, Oliver Hudson and Bianca Kajlich.Technically Russell and Jeff are the main characters of the show, which make viewers wanna watch the show. Russell is a playboy and Jeff is a kind of frustrated family man, The relationship wiz... with an experience of all the problems a married couple face in a relationship.Patrick Warburton - does a {\color{ao}{\bf horrible}} ({\color{amethyst} {\bf awful}}) job as Jeff, he is not at all suited for the role. He is like a robot, literally there is no punch in his dialog delivery. Cast is really very important for viewers to like it. The bad acting certainly will take the show downhill...} \\\bottomrule
\end{tabular}
\caption{Illustrations of generating adversarial training examples to flip the label of a test example.  {\color{red}{\bf Red}} in the test example denotes the salient word for the golden label. 
{\color{ao}{\bf green}} in the test example denotes the salient word for an incorrect label.  For the training example with golden label, the {\color{red}{\bf red}}  marked token and its following {\color{amethyst} purple} token are respectively the original text and the
substitution in the adversarial example for $\downarrow I(z,w_y(\mathbf{e}))$. 
For the training example with the incorrect label, the {\color{ao}{\bf green}}  marked token and its following {\color{amethyst} purple} token are respectively the original text and the
substitution in the adversarial example for   $\uparrow I(z,w_{y'}(\mathbf{e}))$.
}
\label{example}
\end{table*}

\subsection{Generating Adversarial Examples}
There has been a growing interest in performing adversarial attacks against an existing neural model, both in vision \citep{goodfellow2014explaining,kurakin2016adversarial,athalye2018synthesizing}  and NLP \citep{ebrahimi2017hotflip,wang2020fast,Zhu2020FreeLB,zhou2020defense}.
Here we describe how the proposed paradigm can be used  for this purpose.
There are two directions that we can take to generate adversarial examples:

(a) 
$\downarrow I(z,w_y(\mathbf{e}))$: 
for a text example $z = (x,y)$, where $x$ is the input and $y$ is the gold label. 
$w_y(\mathbf{e})$ denotes the  score of $w_y(\mathbf{e})$ for the top salient word(s) $x_t$ with embedding $\mathbf{e}$.  We can modify  
the
point ${z}$ to $\tilde{z}$
 to  decrease $w_y(\mathbf{e})$, making the model not focus on the region that it should focus on. For this purpose, $\tilde{z}$ should be as follows:
\begin{equation}
\tilde{z} =\arg\max_{\tilde{z}} I(z, \tilde{z}, w_y(\mathbf{e}) )
\label{max}
\end{equation}
(b) 
  $\uparrow I(z,w_{y'}(\mathbf{e}))$: 
 for an incorrect label $y'\neq y$, associated with the saliency  score of $w_{y'}(\mathbf{e})$ for top salient regions $\mathbf{e}$,  we can modify the training point  ${z}$
 to $\tilde{z}$
 to most increase $w_{y'}(\mathbf{e})$, making the model  focus on the region that it should not focus on. $\tilde{z}$ should be as follows:
\begin{equation}
\tilde{z} =\arg\min_{\tilde{z}} I(z,\tilde{z}, w_{y'}(\mathbf{e}))
\label{min}
\end{equation}
(c) Combined: combining (a) and (b).

(a) and (b) can be readily applied in  computer vision due to its continuous nature of images, but need further modifications in NLP since words are discrete. 
We follow general protocols for word substitution in generating adversarial sentences in NLP \cite{ren2019generating}, in which 
a word $x_t$ can only be replaced by its synonym found in the WordNet.
The synonym list for $x_t$ is
 denoted by $L(x_t)$. 
For a given training example $\mathbf{x} = \{x_1,x_2,...,x_N\}$, 
it can be changed to $\tilde{\mathbf{x}}$ by replacing $x_t$ with $\tilde{x}_t \in L(x_t)$.

\begin{table}[t]
\centering
\small
\begin{tabular}{cccc}\toprule
 Setting& \# Examples & IMDB$\%$   & AGNews$\%$  \\\midrule
{Origin} & --- & 93.2  &94.4 \\\hline
$\downarrow I(z,w_y(\mathbf{e}))$ & 1 &  68.1 &76.8\\
$\downarrow I(z,w_y(\mathbf{e}))$ & 2 &  52.0&63.3 \\
$\downarrow I(z,w_y(\mathbf{e}))$ & 6 &  19.7  &36.2\\
$\downarrow I(z,w_y(\mathbf{e}))$ & 10 &  2.1&18.4\\
$\downarrow I(z,w_y(\mathbf{e}))$ & 20 &  0.0&6.9\\\hline
$\uparrow I(z,w_{y'}(\mathbf{e}))$ & 1 &75.4& 82.0\\
$\uparrow I(z,w_{y'}(\mathbf{e}))$ & 2 &60.5&71.5\\
$\uparrow I(z,w_{y'}(\mathbf{e}))$ & 6 &35.3&49.2\\
$\uparrow I(z,w_{y'}(\mathbf{e}))$ & 10& 8.6&30.6\\
$\uparrow I(z,w_{y'}(\mathbf{e}))$ & 20& 0.2&10.3\\\hline
{Combined} & 2 &  48.7& 57.5\\
{Combined} & 6 & 18.4&34.1\\
{Combined} &  10 & 1.8& 16.5\\
{Combined} &  20 & 0.0&3.9 \\
\bottomrule
\end{tabular}
\caption{Results of adversarial attacks on the IMDB dataset. ``\# Examples'' denotes the number of perturbed training examples for {\it each} test example.}
\label{adver}
\end{table}

\paragraph{Efficient Implementation} Suppose that there are $N$ words in $\mathbf{x}$, and the average size of $L(x_t)$ is $|L(x_t)|$. We need to enumerate all $|L(x_t)| \times N$ potential $\tilde{z}$ to obtain the optimal in Eq.\ref{max} and Eq.\ref{min}. This is  computationally intensive. 
Towards efficient computation, we decouple the process into two stages, in which we first select the optimal token $x_t \in x$  which is most contributive to 
 the salient region: 
\begin{equation}
t = \arg\max_{t \in [1, N]}   I(x_t, w_y(\mathbf{e}) )
\end{equation}
Next, by fixing $x_t$, we iterate over its synonyms and obtain $\tilde{z}$ with largest score of Eq.\ref{max}.
\begin{equation}
\tilde{z} =\argmax_{\tilde{z}= z-x_t + \tilde{x}_t : \tilde{x}_t \in L(x_t)} I(z, \tilde{z}, w_y(\mathbf{e}) )
\end{equation}
This two-stage strategy is akin to the greedy model taken in  \newcite{ren2019generating}. 
To avoid the local optimal, we also randomly sample $T=5$ tokens as candidates for $x_t$ and use the synonyms of 
each candidate
 to obtain the values of Eq.\ref{max}.  
The scores is compared with the output  score from the two-stage process, and  the best one is remained. 
Similar strategy is applied to Eq.\ref{min}. An example is shown in Table \ref{example}.

After iterating over all test examples, 
we can obtain a new set of training examples. 
The model is retrained on the newly generated training examples.  
This iterating strategy is similar to training-set analogue of the
methods adopted in \newcite{koh2017understanding,goodfellow2014explaining}.
It is worth noting that the proposed paradigm for  adversarial attacks is fundamentally different from the group of work for performing adversarial attacks in NLP \cite{ren2019generating, alzantot2018generating, zhang2020generating} in that the proposed method changes training examples while the rest focus on the change of test examples. 

We used two datasets for test, IMDB and AG’s News. 
Due to the computational intensity, 
We use BERT as the model backbone. 
Results are shown in Table \ref{adver}. Simply perturbing one training example for each test example has a significant negative impact on the prediction performances. For
 IMDB,
forcing the model not to focus on the salient regions ($\downarrow$) leads to an accuracy drop by nearly 25\%, and forcing the model to focus on unimportant regions ($\uparrow$) leads to a drop by about 18\%. Continuously increasing the number of perturbed training examples proceeds to deteriorate the performance. For example, increasing one more perturbed training example makes the model only obtain an accuracy of 52.0\%, 60.5\% and 48.7\%, respectively based on the method of  $\downarrow I(z,w_y(\mathbf{e}))$, $\uparrow I(z,w_{y'}(\mathbf{e}))$ and {Combined}. As we increase the number of perturbed training examples to 20, the model almost can not correctly predict any test example. 
We also observe  
 that with the same number of  perturbed training examples, the $\downarrow I(z,w_y(\mathbf{e}))$ method always outperforms $\uparrow I(z,w_{y'}(\mathbf{e}))$. This is  because shifting the model's ``correct'' focus (salient regions) to other ``incorrect'' parts (non-salient regions) is easier to hinder the model from making right predictions than pushing the model to focus on ``incorrect'' parts.
Another interesting observation is that the {\it combined} setting outperforms both $\downarrow I(z,w_y(\mathbf{e}))$  and  $\uparrow I(z,w_{y'}(\mathbf{e}))$ with the same number of
perturbed training examples. This is because the {\it combined} setting takes the advantage of both methods, and avoid the duplicates or overlaps in semantics for different perturbed training examples.   
Similar trends are observed for the AG's News dataset, and are thus omitted for brevity. 

\subsection{Fixing Model Predictions}
The proposed model can also be used to make changes to training examples to fix model predictions. 
This can be done by doing exactly the opposite of methods taken for  generating  adversarial examples:
(a) increasing $I(z,w_y(\mathbf{e}))$, denoted by $\uparrow I(z,w_y(\mathbf{e}))$, generating training examples to make the model focus on the region that it should focus on; and 
(b) decreasing $I(z,w_{y'}(\mathbf{e}))$, denoted by $\downarrow I(z,w_{y'}(\mathbf{e}))$, generating training examples to make the model not focus on the region that it should not focus on. Since our goal is to generate training examples to fix model's false predictions, 
we don't have to be bonded by the principle in adversarial example generation that the perturbation is (nearly) human indistinguishable. 
$L(x_t)$ is thus set to both synonyms and antonyms of $x_t$. 

\begin{figure}
    \centering
    \includegraphics[scale=0.5]{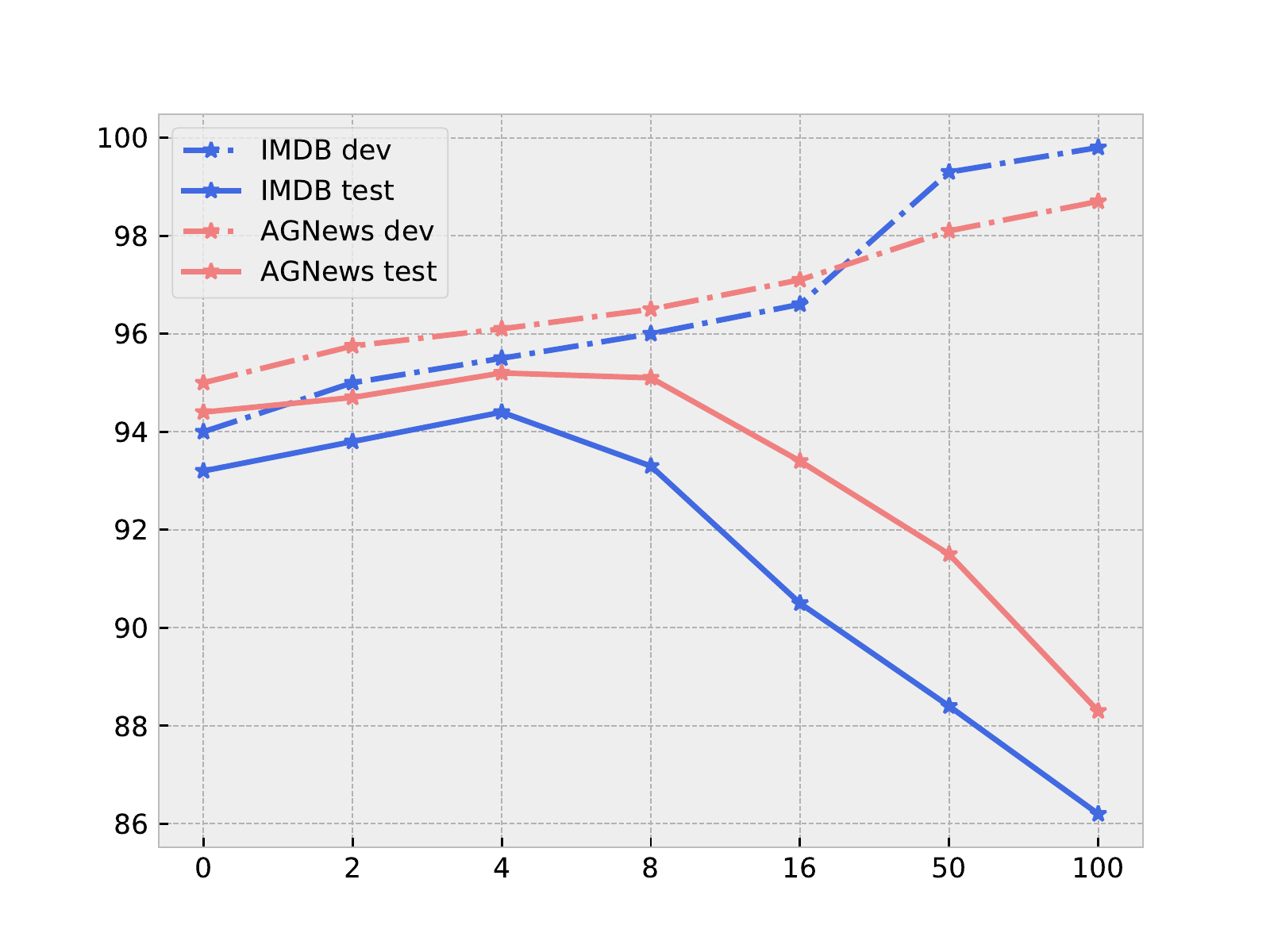}
    \caption{Performances on dev and test sets w.r.t. the number of generated training examples for each misclassified test example.}
    \label{lineplot}
\end{figure}

Since we don't have access to the test set, we use the dev set to perturb training examples. 
For each erroneously classified example in the dev set, we generate perturbed training examples, trying to flip the decision. 
Instead of replacing training examples,
perturbations are additionally added to the training set.
With the combined training set, a model is retrained and we report test performances.  
Results are shown in Figure \ref{lineplot}. 
As we can see, the trends for both IMDB and AG News are the same: when we gradually increase the number of training examples generated for each misclassified test example from 0 to 8, the test performance increases accordingly. But as we continue increasing training examples, 
the model trained on the generated training examples starts overfitting the dev set, leading to 
drastic  drops in
test performances. We obtain the best test results at \# Examples = 4, giving a +1 accuracy gain compared to models without augmentation.

\section{Conclusion}
In this paper, we propose a paradigm that unifies test-focused methods and training-focuses for interpreting a neural network model's prediction.
The proposed paradigm answers the question of 
{\it because of  
which part(s) of
which training example(s), the model attends to which part(s) of a given test example}. 
We demonstrate that the proposed methodology  offers clear
explanations about neural model decisions, along with being useful for error analysis and  creating adversarial examples.

\bibliography{emnlp2020}
\bibliographystyle{acl_natbib}

\appendix
\section{Derivation of The Influence Function}
\label{appendixA}
In this section, we follow \citet{koh2017understanding} to prove the following expression (Eq.\ref{eq3}):
\begin{equation}
  I(z,{\theta})\triangleq\frac{\mathrm{d}{\theta}_{z,\epsilon}}{\mathrm{d}\epsilon}\Big|_{\epsilon=0}=-H^{-1}_{{\theta}}\nabla_\theta L(z;{\theta})
\end{equation}
Recall that ${\theta}$ minimizes the empirical risk $R(\theta)$:
\begin{equation}
  R(\theta)\triangleq\frac{1}{n}\sum_{i=1}^nL(z_i;\theta)
\end{equation}
and $\theta_{z,\epsilon}$ minimizes the following upweighted empirical risk:
\begin{equation}
    R(\theta)+\epsilon L(z;\theta)
\end{equation}
Therefore, we have the following identities hold:
\begin{equation}
\begin{aligned}
    \nabla R(\theta)&=0\\
    \nabla R(\theta_{z,\epsilon})+\nabla \epsilon L(z;\theta_{z,\epsilon})&=0
\end{aligned}
\end{equation}
for which we can apply first-order Taylor expansion to the second expression:
\begin{equation}
  \begin{aligned}
    0\approx&\left[\nabla R({\theta})+\epsilon\nabla L(z;{\theta})\right]\\
    &\left[\nabla^2 R({\theta})+\epsilon\nabla^2 L(z;{\theta})\right]\Delta_\epsilon
  \end{aligned}
\end{equation}
where $\Delta_{\epsilon}={\theta}_{z,\epsilon}-{\theta}$ is the parameter change. Solving for $\Delta_\epsilon$, we have:
\begin{equation}
  \begin{aligned}
    \Delta_\epsilon\approx&-\left[\nabla^2 R({\theta})+\epsilon\nabla^2 L(z;{\theta})\right]^{-1}\\
    &\left[\nabla R({\theta})+\epsilon\nabla L(z;{\theta})\right]
  \end{aligned}
\end{equation}
Dropping higher order terms of $\epsilon$ and plugging $\nabla R(\theta)=0$ in, we have:
\begin{equation}
  \Delta_\epsilon\approx -\nabla^2R({\theta})^{-1}\nabla L(z;{\theta})\epsilon
\end{equation}
which shows that
\begin{equation}
  \frac{\mathrm{d}{\theta}_{z,\epsilon}}{\mathrm{d}\epsilon}=\frac{\mathrm{d}\Delta_\epsilon}{\mathrm{d}\epsilon}\Big|_{\epsilon=0}=-H^{-1}_{{\theta}}\nabla_{\theta}L(z;{\theta})
\end{equation}

\section{Efficient Computation of The Hessian}
Albeit its appealing properties for approximating influences, calculating the full Hessian $H_{{\theta}}$ is too expensive to afford. To efficiently calculate it, we follow \citet{koh2017understanding} to use the method of stochastic estimation proposed by \citet{agarwal2016second} and the Hessian-Vector Product (HVP) trick \citep{10.1162/neco.1994.6.1.147} to get an estimator that only samples a single point per iteration, leading to a trade-off between speed and accuracy. 

Concretely, we let $H^{-1}_j\triangleq\sum_{i=0}^j(I-H)^i$ be the first $j$ terms in the matrix Taylor expansion of $H^{-1}$, which can be recursively written as $H_j^{-1}=I+(I-H)H^{-1}_{j-1}$. It can be validated that $H^{-1}_j\to H^{-1}$ as $j\to\infty$. So we can replace $H$ with a single drawn point from the training set, and iteratively recover $H^{-1}$.

In particular, we first uniformly sample $t$ points $z_{s_1},\cdots,z_{s_t}$ from the training data, and define $\tilde{H}_0^{-1}v=v$. Then we recursively compute $\tilde{H}_j^{-1}v=v+(I-\nabla_\theta^2 L(z_{s_j};{\theta}))\tilde{H}^{-1}_{j-1}v$ according to the above expression. After $t$ iterations, we take $\tilde{H}^{-1}_tv$ as the final unbiased estimate of $H^{-1}v$. We pick $t$ such that $\tilde{H}_t$ stabilizes. To further reduce variance we repeat this process $r$ times and average the results. Afterwards, we apply the process to calculate the Hessian-vector products in $I(z,\theta), I(z,w_y(\mathbf{e}))$ and $I(z,\tilde{z},\theta), I(z,\tilde{z},w_y(\mathbf{e}))$.

\end{document}